\documentclass[10pt,twocolumn,letterpaper]{article}

\usepackage{cvpr}              % To produce the CAMERA-READY version
\usepackage{mmstyle}
\usepackage{comment}
\usepackage{multirow}
\usepackage{bbding}
\usepackage{indentfirst}
\usepackage[dvipsnames]{xcolor}

\definecolor{cvprblue}{rgb}{0.21,0.49,0.74}
\usepackage[pagebackref,breaklinks,colorlinks,citecolor=cvprblue]{hyperref}

\def\paperID{11} % *** Enter the Paper ID here
\def\confName{CVPR}
\def\confYear{2024}

\title{ReasonPix2Pix: Instruction Reasoning Dataset for Advanced Image Editing}

\author{%
Ying Jin$^{1}$ \quad Pengyang Ling$^{2}$ \quad Xiaoyi Dong$^{2}$ \quad Pan Zhang$^{2}$  \quad Jiaqi Wang$^{2} {\textsuperscript{\Envelope}}$\thanks{\textsuperscript{\Envelope}Corresponding author.} \quad Dahua Lin$^{1,2}$ \vspace{5pt} \\
$^1$CUHK-SenseTime Joint Lab, The Chinese University of Hong Kong \\
$^2$Shanghai AI Laboratory \\
{\tt\small \{jy021,dhlin\}@ie.cuhk.edu.hk, lpyang27@mail.ustc.edu.cn},\\
{\tt\small\{dongxiaoyi, zhangpan, wangjiaqi\}@pjlab.org.cn}
\\
}

\begin{document}

\twocolumn[{%
\renewcommand\twocolumn[1][]{#1}%
\maketitle
    \begin{center}
        \includegraphics[width=1\linewidth]{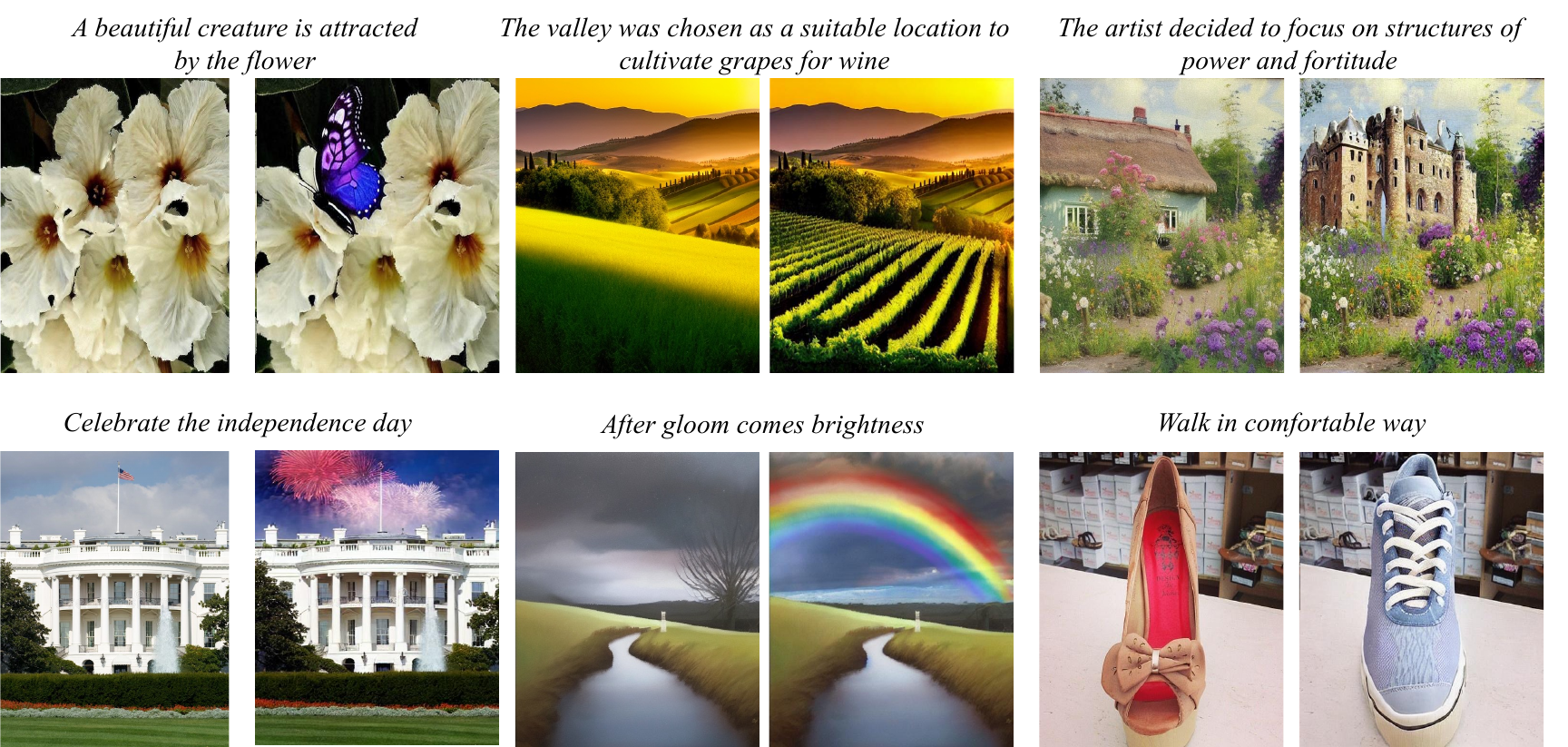}
        \vspace{-2.3mm}
        % \captionof{figure}{We introduce ReasonPix2Pix, a more intelligent image editing framework with reasoning ability. Given an implicit instruction, ReasonPix2Pix can understand the instruction and then produce an appropriate edited image. }
        \captionof{figure}{Generated results from the model trained on our dataset. Given an implicit instruction, our model can understand the instruction and then produce an appropriate edited image. }
        \label{fig:teaser}
        \vspace*{2.7mm}
    \end{center}%
}]

\begin{abstract}
% Instruction-based image editing focuses on equipping a generative model with the capacity to adhere to human-written instructions for editing images. Current approaches typically comprehend explicit and specific instructions. However, they often exhibit a deficiency in executing active reasoning capacities required to comprehend instructions that are implicit or insufficiently defined. To enhance active reasoning capabilities and impart intelligence to the editing model, we introduce ReasonPix2Pix. This is a Multimodal Large Language Model-assisted instruction editing framework. To stimulate the reasoning ability of the model  for image editing, we develope a comprehensive reasoning-attentive instruction editing dataset.  When fine-tuned with the proposed dataset under supervised conditions, ReasonPix2Pix demonstrates superior performance in instructional editing tasks, independent of whether the tasks require reasoning or not. The code, model, and dataset will be publicly available.

Instruction-based image editing focuses on equipping a generative model with the capacity to adhere to human-written instructions for editing images. Current approaches typically comprehend explicit and specific instructions. However, they often exhibit a deficiency in executing active reasoning capacities required to comprehend instructions that are implicit or insufficiently defined. To enhance active reasoning capabilities and impart intelligence to the editing model, we introduce ReasonPix2Pix, a comprehensive reasoning-attentive instruction editing dataset. The dataset is characterized by 1) reasoning instruction, 2) more realistic images from fine-grained categories, and 3) increased variances between input and edited images. When fine-tuned with our dataset under supervised conditions, the model demonstrates superior performance in instructional editing tasks, independent of whether the tasks require reasoning or not. The code will be available at \url{https://github.com/Jin-Ying/ReasonPix2Pix}.
\end{abstract}

\section{Introduction}
\label{sec:intro}
%Instruction-based image editing aims to teach a generative model to follow human-written instructions for image editing. It makes the AIGC system to understand the explicit and specific instruction of human, such as \textit{replace the fruits with cake}. However, current instruction-based image editing models lack reasoning ability to understand or analysis the indirect and vague instructions. For examples, instruction \textit{make the room tidy} should be analyzed and decomposed to many explicit and specific instructions \textit{replace loose clothing with neatly folded clothing, remove trash from the floor, make the bed smooth}, instruction \textit{let her be in ancient times} should be analyzed and decomposed to instructions \textit{change the background to an ancient style, replace her costume with ancient costume}. Such self-reasoning ability helps the AIGC system understand the intent of the huaman and is crucial in developing next-generation intelligent AIGC systems
Instruction-based image editing aims at furnishing a generative model with the capacity to adhere to human-written instructions for editing images, which is vital to facilitate the AI-Generated Content (AIGC) system’s comprehension of human intentions.  

\begin{figure}[h!]
    \begin{subfigure}[t]{0.327\linewidth}
        \includegraphics[width=\linewidth]{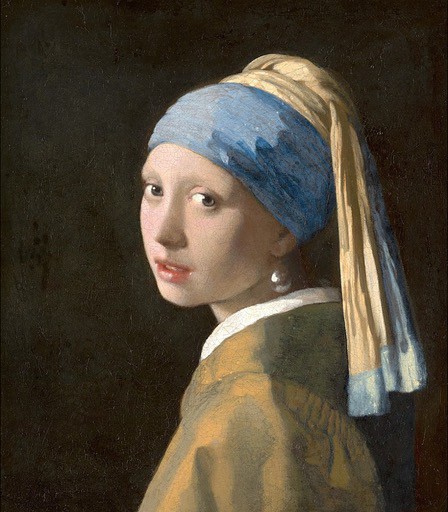}
        \begin{center}
        \vspace{-2.5mm}
        \scriptsize Input
        \vspace{-1.5mm}
        \end{center}
    \end{subfigure}%
    \hfill%
    \begin{subfigure}[t]{0.327\linewidth}
        \includegraphics[width=\linewidth]{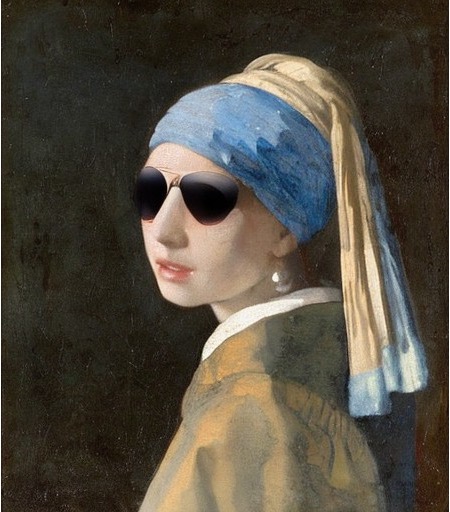}
        \begin{center}
        \vspace{-2.5mm}
        \scriptsize Add a pair of sunglasses
        \vspace{-1.5mm}
        \end{center}
    \end{subfigure}%
    \hfill%
    \begin{subfigure}[t]{0.327\linewidth}
        \includegraphics[width=\linewidth]{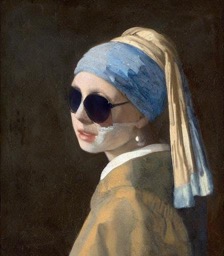}
        \begin{center}
        \vspace{-2.5mm}
        \scriptsize She prefers face mask to sunglasses
        \vspace{-1.5mm}
        \end{center}
    \end{subfigure}%
    
    \caption{Previous method, InstructPix2Pix, is capable of tackling instruction \textit{``add a pair of sunglasses"}, but it generates absolutely wrong result for the instruction \textit{``she prefers face mask to sunglasses"}.}
    \label{fig:pearl-mistake}
\end{figure}

\begin{figure*}[pt]
   \begin{center}
   \includegraphics[width=1\linewidth]{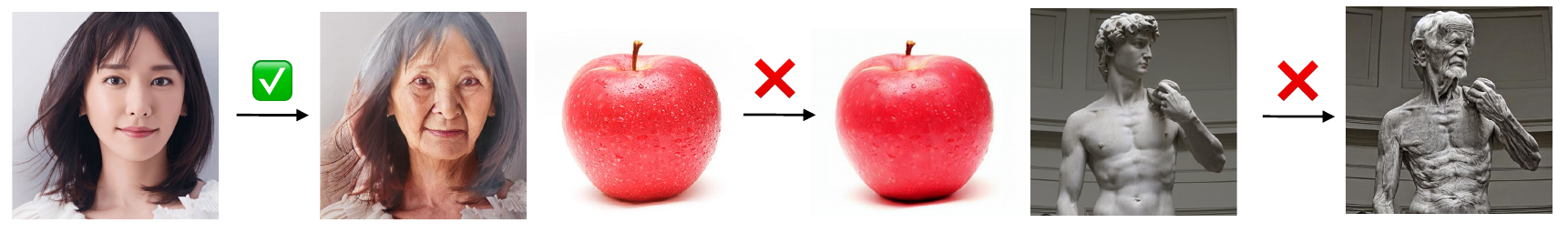}
   \end{center}
   \vspace{-10pt}
   \caption{For the instruction \textit{``make it 50 years later"}, previous methods can make a young woman an old one, but cannot generate any results for fruit input (apple). In addition, when the input is a statue man, previous methods still make it old, which is wrong. The reasonable results may be an old woman, a rotted fruit, and a broken statue respectively. Therefore, these methods lack the capability of comprehending images with instruction.}
   \label{fig:50-years}
\end{figure*}

Prevailing instruction-based image editing frameworks typically comprehend explicit and specific instructions, such as \textit{``replace the fruits with cake"}. 
Unfortunately, these models display a deficiency in active reasoning capabilities, \ie understanding the instructions rather than extracting words from them. As shown in Figure~\ref{fig:pearl-mistake}, one typical instruction-based image editing framework, InstructPix2Pix, fails to realize \textit{``she prefers face mask to sunglasses"}, adding sunglasses to the woman, which is unreasonable. Meanwhile, the model lacks the ability to comprehend the given image. 
For example, for a simple straightforward instruction \textit{``make it 50 years later"}, with a variety of given images, the editing outcomes should be different. But in Figure~\ref{fig:50-years}, previous methods simply turn the person to an older one or even fail to edit the image, which is absolutely incorrect. 

\begin{table*}[tp]
\centering
\caption{Comparison of different datasets. }
\resizebox{0.95\textwidth}{!}{
\begin{tabular}{l|cc|c}
\toprule
Dataset & Real Image & Reasoning Instruction & Text Example  \\
\midrule
InstructPix2Pix~\cite{brooks2023instructpix2pix} & \XSolidBrush  & \XSolidBrush & \textit{Turn fruits to a cake}\\
MagicBrush~\cite{zhang2023magicbrush} & \Checkmark & \XSolidBrush  & \textit{Make him wear a hat}\\
ReasonPix2Pix (Ours) & \Checkmark & \Checkmark & \textit{The beautiful woman wants to walk in a more comfortable manner} \\
\bottomrule
\end{tabular}
}
\label{tab:data-comp}
\end{table*}

On the other hand, these methods also lack the capacity to comprehend implicit or inadequately defined instructions. This requires manual intervention to either make implicit instructions explicit or deconstruct the instructions into multiple explicit, specific instructions to align with the capabilities of these models. 
For example, the instruction \textit{``make the room tidy"} necessitates manual partitioning into various steps such as \textit{``replace loose clothing with neatly folded clothing"}, \textit{``remove garbage from the floor"}, and \textit{``smooth the bed linens"}, among others. 
Similarly, the implicit instruction \textit{``she is the star of the show"} necessitates human intervention to render it explicit as \textit{``add some sparkles and a spotlight effect to the image"}.
%Similarly, the instruction \textit{``let her be in ancient times"} requires human intervention to prompt these models not to neglect elements like \textit{``replacing her contemporary attire with period-appropriate garments"}, in addition to \textit{``change the background to reflect an ancient era"}.
So, enhancing the self-reasoning capabilities is not only user-friendly, but also key to the advancement of next-generation intelligent AIGC systems.

% Instruction reasoning for image editing necessitates not only the reliance on reasoning capability of LLM, but also demands the ability of comprehending the given image. 
% For instance, when presented with the instruction \textit{``make it 50 years later"} for a variety of given images, the editing outcomes should be different. Accordingly, we employ a Multimodal Large Language Model (MLLM) instead of a singular large language model.

The potential of generative models aided by MLLM for reasoning-aware instruction editing is considerable. Nonetheless, existing datasets designated for instruction editing fail to fully unlock and exploit the inherent reasoning capabilities of the models. Thus,  we develop a comprehensive reasoning-attentive instruction editing dataset, ReasonPix2Pix, comprising both image pairs and accompanying reasoning instructions. ReasonPix2Pix is characterized by: 1) implicit instructions to further the model's reasoning capability, 2) an abundance of real images taken from fine-grained categories, and 3) increased variances between the input and the subsequently edited images, particularly at the geometric level. We compare it with previous datasets in Table~\ref{tab:data-comp}.

Further, we inject MLLM into the image editing model and fine-tune it on our dataset, which enhances the reasoning abilities of image editing and significantly boosts the quality of instructional editing. Our contribution can be summarized as follows:
\begin{itemize}[leftmargin=*]
\vspace{0.3em}
\itemsep0.3em
% \item We propose ReasonPix2Pix, an instruction-based image editing framework with reasoning capabilities. Leveraging the power of the MLLM, ReasonPix2Pix demonstrates its intelligence in understanding human intent.
\item We propose image editing with instruction reasoning, an interesting task to enhance the model's intelligence in understanding human intent.
\item We develop a comprehensive reasoning-attentive instruction editing dataset, ReasonPix2Pix. It comprises both image pairs and accompanying reasoning instructions.
\item We fine-tune a simple framework on our dataset. Without hustle and bustle, our model demonstrates not only superior performance in instruction editing tasks that do not necessitate reasoning but also performs adequately in tasks necessitating reasoning.
\end{itemize}

\section{Related Work}
\label{sec:related}
\paragraph{Image Editing}
Image editing is a fundamental computer vision task, which can also be viewed as an image-to-image translation. Numerous works~\cite{gatys2015neural,gatys2016image,huang2018munit,liu2019few,ojha2021few-shot-gan} have been invented to tackle this task, especially after the proposal of Generative Adversarial Networks (GAN)~\cite{pix2pix2017,CycleGAN2017,huang2018munit}. One line of methods~\cite{abdal2019image2stylegan,abdal2020image2stylegan++,alaluf2022hyperstyle,epstein2022blobgan,tov2021designing,richardson2021encoding,chai2021latent} plug the raw image into latent space~\cite{karras2019style,karras2020analyzing}, and them manipulate it. These methods are proven to be effective in converting image style, adding and moving objects in the images. Recently, with the explosion of multi-modal learning, text information can be embedded through models such as CLIP, and then serves as guidance for image editing~\cite{kwon2022clipstyler,Patashnik_2021_ICCV,zheng2022bridging,gal2022stylegan,nichol2021glide,kim2022diffusionclip,avrahami2022blended, crowson2022vqgan, bar2022text2live}. These methods enable models to edit images according to the given text. 

\paragraph{Diffusion Models} Diffusion model~\cite{pmlrv37sohldickstein15} is among the most popular generative models, showing strong performance in image synthesis~\cite{song2019generative,ho2020denoising,dhariwal2021diffusion,ho2022cascaded,saharia2022image,saharia2022palette,zhou2024nodi}. It learns the probability distribution of a given dataset through a diffusion process. Recently, text-to-image diffusion models~\cite{nichol2021glide,rombach2022high,ramesh2022hierarchical,saharia2022photorealistic}, such as Stable Diffusion~\cite{rombach2022high}, achieve great success in converting text to high-quality images. 

\paragraph{Diffusion Models for Image Editing} Some of the diffusion models are naturally capable of editing images~\cite{meng2021sdedit,kawar2022imagic,avrahami2022blended,ramesh2022hierarchical,hertz2022prompt}. However, when applied to practice, these models show weak stability (\ie, generating similar images when given similar texts). This problem is alleviated by imposing constrains on models~\cite{hertz2022prompt} via Prompt-to-Prompt. Different from previous methods that tackle generated images, SDEdit~\cite{meng2021sdedit} edits real images by a noising-and-denoising procedure. 

Image inpainting can be viewed as a more refined image editing. It converts text inputs and user-drawn masks~\cite{ramesh2022hierarchical,avrahami2022blended,kawar2022imagic} to images of a specific category or style by learning from a small set of training samples~\cite{gal2022image,ruiz2022dreambooth}. InstructPix2Pix~\cite{brooks2023instructpix2pix} simplify the generation procedure, taking one input image and one instruction to edit it without any training. It proposes a large-scale dataset, with paired images and the corresponding instruction. However, it only contains straightforward instructions, which hinders it from being applied to complicated real-world scenarios. So in this paper, we construct the instruction reasoning dataset for improving image editing. 

\paragraph{Multi-modal Large Language Model} With the rapid development of Large Language Models (LLM), they are extended to more modalities (\eg vision), forming multi-modal large language models. BLIP-2~\citep{li2023blip} and mPLUG-OWL~\citep{ye2023mplug} introduce a visual encoder to embed images, and then combine them with text embeddings. Instruct-tuning is widely adopted to transfer the ability in LLM to the visual domain~\cite{li2023otter,liu2023visual,zhu2023minigpt}. Another line of works use prompt engineering~\citep{wu2023visual,yang2023mm,shen2023hugginggpt,liu2023internchat,yang2023gpt4tools}, which sacrifices end-to-end training. The application of multi-modal large language models to vision tasks~\cite{wang2023visionllm} are proven to be effective in grounding~\cite{peng2023kosmos}, and object detection~\cite{detgpt,zhang2023gpt4roi}.

\section{Methodology}
\label{sec:method}

Our objective is to perform image editing in accordance with human instructions, with a particular emphasis on reasoning instructions. Given an input image denoted as $\vx_{\text{input}}$ and a human instruction denoted as $I$, our model is designed to comprehend the explicit or implicit intents of human and subsequently generate the output image $\vx_{\text{edit}}$ that aligns with the provided instruction. 
To achieve this objective, we introduce ReasonPix2Pix (Section~\ref{sec:ReasonPix2Pix}), a dataset specifically tailored for instruction-based image editing with a focus on reasoning capabilities. Taking our dataset as the foundation training data, we fine-tune a simple framework that comprises a multimodal large language model coupled with a diffusion model. 

\subsection{Preliminaries}
\begin{figure}[pt]
   \begin{center}
   \includegraphics[width=1\linewidth]{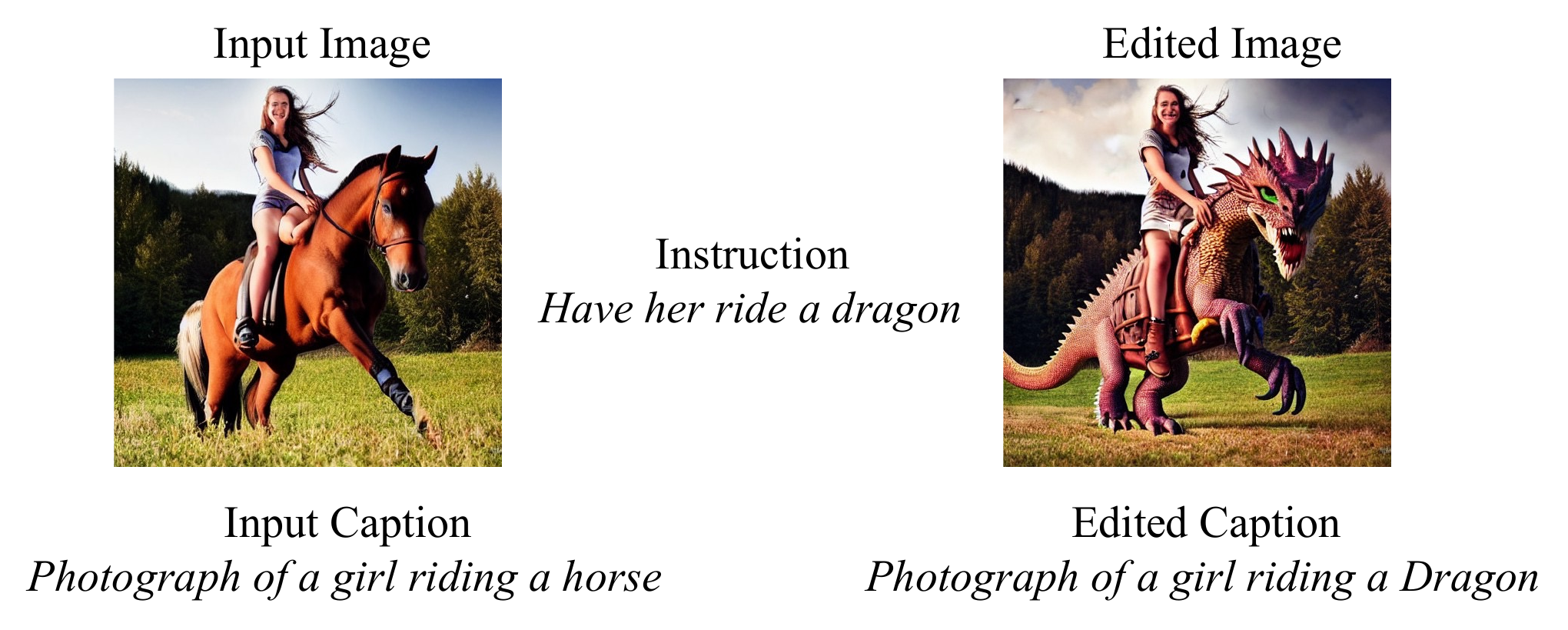}
   \end{center}
   \vspace{-10pt}
   \caption{One sample from InstructPix2Pix dataset. For each paired image, it contains 1) the input image and input caption, 2) the edited image and edited caption, and 3) instructions.}
   \label{fig:p2p-sample}
\end{figure}

\paragraph{InstructPix2Pix Dataset} InstructPix2Pix~\cite{brooks2023instructpix2pix} produces an important large-scale paired dataset to enable instruction-based image editing. Concretely, As shown in Figure~\ref{fig:p2p-sample}, it contains 1) Input Image $\vx_{input}$ and Input Caption $C_{input}$, 2) Edited Image $\vx_{edit}$ and Edited Caption $C_{edit}$, and 3) Text Instruction $I$. 

\paragraph{V3Det Dataset} %\textcolor{blue}{[mllm may be better than V3Det]} 
V3Det~\cite{wang2023v3det} is a vast detection dataset with 13,204 categories, and over $240,000$ images. The images appear to be realistic and complex, developing a more general visual perception system. 

\subsection{ReasonPix2Pix}
\label{sec:ReasonPix2Pix}
%Instruction Editing Dataset with Reasoning

\begin{table}[h]
    \centering
    \caption{Comparison of three parts in our dataset. The data source is listed here. Our generated data is noted as 'Generated'.}
    \resizebox{0.48\textwidth}{!}{
    \begin{tabular}{l|cccc}
    \toprule
    &Input Image & Edited Image & Instruction & Number\\
    \midrule
    Part I& InstructPix2Pix & InstructPix2Pix & Generated & 8,013\\
    Part II & InstructPix2Pix & Generated & Generated & 4,141\\
    Part III & V3Det & Generated & Generated & 28,058\\
    
    \bottomrule
    \end{tabular}
    }
    \label{tab:setup}
\end{table}

Towards injecting reasoning ability into the image editing model, we construct a comprehensive reasoning-attentive instruction editing dataset. 
% We compare it with previous datasets in Table~\ref{tab:data-comp}.

According to the generation procedure, our generated dataset can be divided into three parts. As shown in Table~\ref{tab:setup}, Part I takes the original image pair in InstructPix2Pix, with our generated instruction to enable instruction reasoning, in Part II we start from the input image from InstructPix2Pix, generate our own edited image and instructions, and in Part III, we take more realistic images from V3Det, and generate edited images and instructions. 

\paragraph{Filtering} Though achieving great success in instruction-based image editing, InstructPix2Pix model suffers from various failure cases. One typical failure case is that the model is prone to output the original image, \ie conduct no editing. Delving into the dataset, we observe that a proportion of edited images is highly similar to the input image. Therefore, we need to filter this part of data first, we distinguish them by
\begin{equation}
    ||\vx_{input} - \vx_{edit}||_2^2 < \tau,
\end{equation}
where $\tau$ is the threshold value of the divergence between the input image and the edited image. If $\tau$ is too small, the input image and edited image appear to be too similar, so this pair of data will be abandoned. $\tau$ can be a pre-defined value or rank value (\eg pick out the $10\%$ worst data). 

\paragraph{Part I: Reasoning Instruction generation} For this part of data, we take the input and edited image directly from the original InstructPix2Pix Dataset. To equip the model with the reasoning ability to understand instructions, we need to convert existing instructions $I$ to reasoning instructions $I_{rea}$, which is indirect, but still accurate. We take the large-scale language model, GPT-3.5~\cite{Openai2022ChatGPT} (denoted as $G$) to generate our reasoning instruction. 
\begin{figure*}[htbp]
    \begin{center}
        \includegraphics[width=0.9\linewidth]{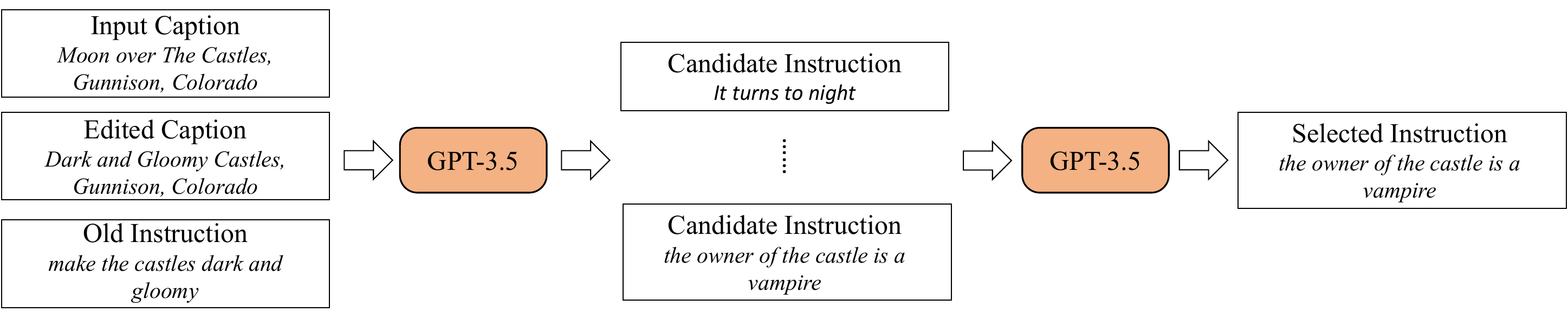}
        
    \end{center}
    
    \caption{Reasoning Instruction Generation. We utilized GPT-3.5 to generate several candidate instructions according to the given input caption, edited caption, and instruction in the original dataset. Then GPT-3.5 selects the best instruction from these candidate instructions.}
    \label{fig:gen}
\end{figure*}

\begin{figure*}[htbp]
    \begin{center}
        \includegraphics[width=0.8\linewidth]{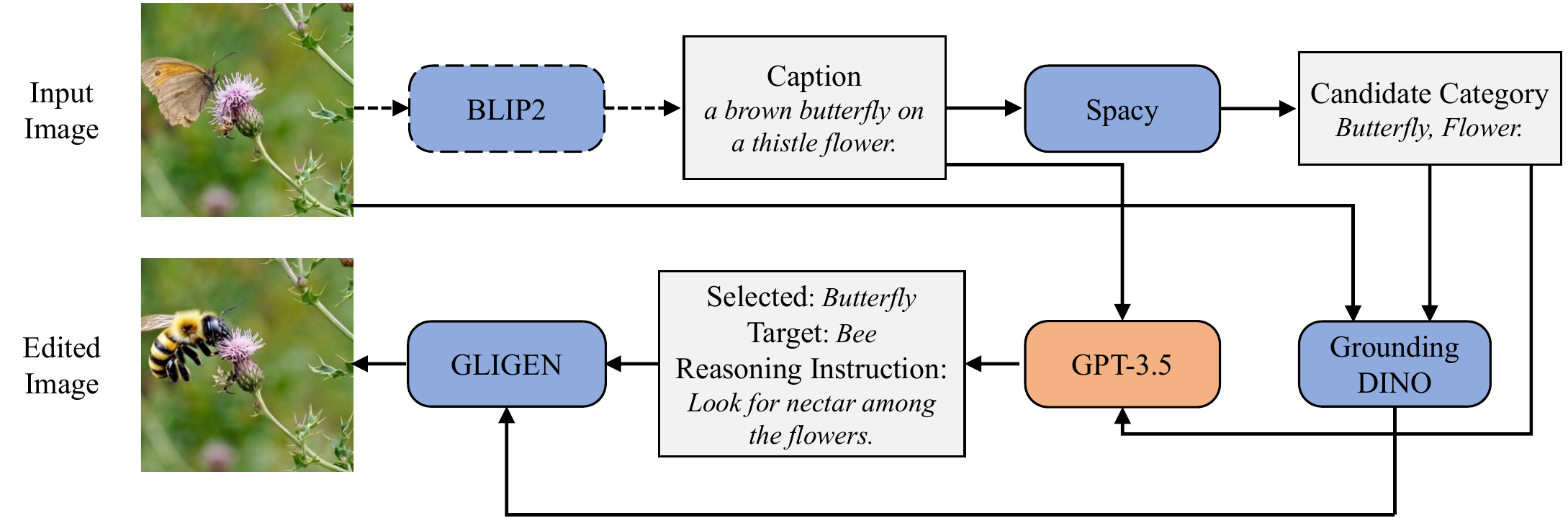}
    \end{center}
    \caption{Image Editing and Reasoning Instruction Generation. If the input image has no caption, BLIP2 will give an image caption. Spacy extracts the candidate categories in the caption. Then, Grounding DINO locates the objects in the image. GPT-3.5 produces 1) selected category, 2) target category, and 3) reasoning instruction. Finally, GLIGEN generates the edited image. }
    \label{fig:gen2}
\end{figure*}

As shown in Figure~\ref{fig:gen}, we take input caption, edited caption, and original instruction from InstructPix2Pix dataset, inject them into GPT-3.5, and ask GPT-3.5 to generate candidate instructions. 
\begin{equation}
    I_{gen} = G(P_{gen}(C_{input}, C_{edit}, I)),
\end{equation}
where $P_{gen}()$ is the generation prompt we design to project the input caption, edited caption, and original instruction to a new instruction $I_{gen}$. With our prompt, the generated instruction should be 1) indirect, and 2) in a similar effect to the original instructions. For each input, we repeat this procedure for $n$ times to generate multiple candidate instructions $I_{gen}^0, ..., I_{gen}^n$. 

Then we again ask GPT-3.5 to distinguish the best instruction among them. 
\begin{equation}
    I_{rea} = G(P_{select}(I_{gen}^0, ..., I_{gen}^n)),
\end{equation}
where $P_{select}()$ is the selection prompt we design to ask the GPT model to select the most suitable instruction from $I_{gen}^0, ..., I_{gen}^n$. The selected instruction is $I_{rea}$ and it will serve as the reasoning instruction in Part I data. Therefore, the Part I data includes $\vx_{input}$, $\vx_{edit}$, and 3) $I_{rea}$. 
% The selected instruction serves as the new instruction in our dataset. 

\paragraph{Part II $\&$ III: Image Editing and Reasoning Instruction Generation} To further improve the model capability, we expand the dataset with the other two parts of data. % These parts of data aim at enhancing the model's ability to tackle more complex object replacement tasks and more realistic images. 
These parts of data not only enhance the reasoning ability of our model, but also aim at improving our model's capability of coping with more realistic images with fine-grained categories and more variances between the input and edited image. 

We take the original input images from InstructPix2Pix and V3Det. The former is the same with Part I but the latter contains abundant real images from large amounts of categories. As shown in Figure~\ref{fig:gen2}, since the images from V3Det have no captions, we first generate a caption through BLIP2. For images from InstructPix2Pix, we take the original input caption directly. For simplicity, we denote the input image and caption as $\vx_{input}$ and $C_{input}$.

The caption is passed to a Spacy model $S$, an advanced Natural Language Processing (NLP) model to recognize entities in sentences. Here we utilize it to extract the candidate categories,
\begin{equation}
    y_1, ..., y_i = S(C_{input}),
\end{equation}
where $S$ extracts $i$ candidate categories. For instance, in Figure~\ref{fig:gen2}, Spacy takes $i=2$ categories,
\ie butterfly and flower. 

With these categories, we can locate the corresponding object in the image by Grounding DINO~\cite{liu2023grounding}. 
\begin{equation}
    b_k = DINO(\vx_{input}, y_k),
\end{equation}
where $b_k$ is the bounding box of object belongs to category $y_k$ in image $\vx_{input}$.

Then we inject the caption and candidate categories into GPT-3.5. Here we design another prompt to ask GPT-3.5 to output 1) one selected category, 2) the target category we need to replace it with, and 3) the reasoning instruction.
\begin{equation}
    y_{select}, y', I_{rea} = G(P_{replace}(C_{input}, y_1, ..., y_i)),
\end{equation}
where $P_{replace}$ is the prompt we introduce. It takes input caption and candidate categories and produces the selected category $y_{select}$, the target category $y'$, and the reasoning instruction. We will replace $y_{select}$ with $y'$.
For Figure~\ref{fig:gen2}, GPT selects butterfly and produces the target category bee, therefore, the butterfly will be transformed into a bee. 

Finally, GLIGEN will conduct object replacement from $y_{select}$ category to $y'$ category.
\begin{equation}
    \vx'_{edit} = GLIGEN(\vx_{input}, y_{select}, b_{select}, y'),
\end{equation}
where $\vx'_{edit}$ is the generated edited image, and $b_{select}$ is the bounding box of the object belongs to $y_{select}$. 
Through this procedure, we obtain paired image ($\vx_{input}$ and $\vx'_{edit}$) and instruction $I_{rea}$, forming Part II and Part III in our dataset. 

\begin{figure}[pt]
   \begin{center}
   \includegraphics[width=1\linewidth]{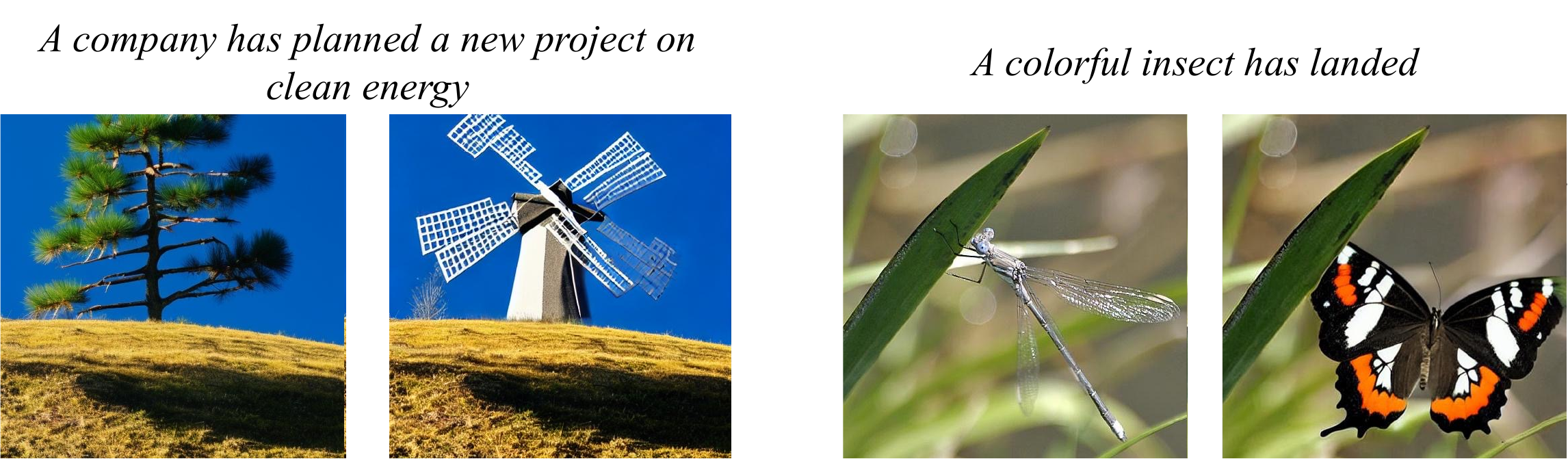}
   \end{center}
   \caption{Data Sample. Our data has 1) reasoning instruction, 2) more variances between input and edited image, and 3) more realistic images.}
   \label{fig:our-data-sample}
\end{figure}

We present some samples of our data here. From Figure~\ref{fig:our-data-sample}, our dataset has complicated reasoning instructions (\eg \textit{``A company has planned a new project on clean energy"}), more changes between input and edited image, especially on geometric level, and more realistic images. 

\begin{figure}[pt]
   \begin{center}
   \includegraphics[width=1\linewidth]{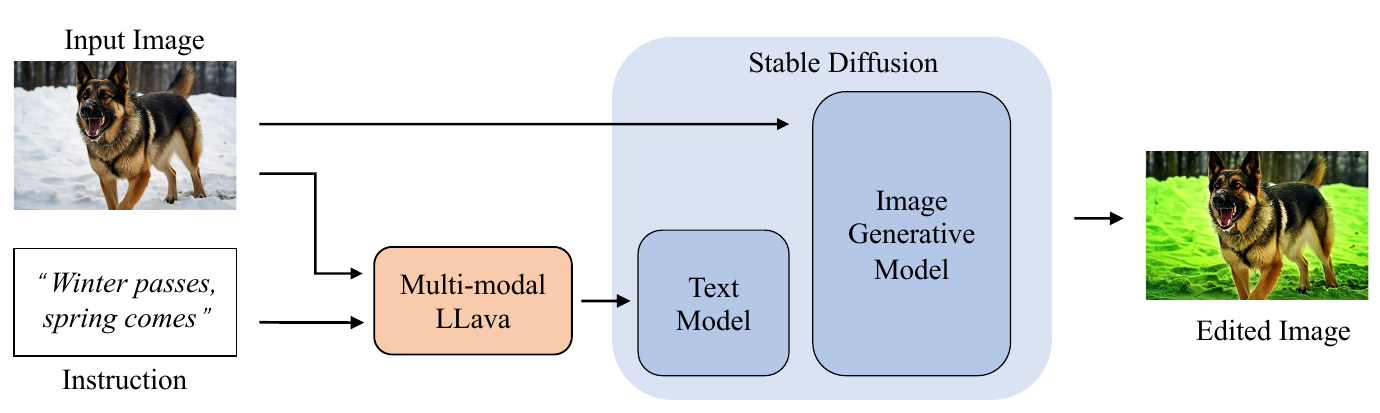}
   \end{center}
   \caption{Method Overview. Multi-modal LLava takes both input image and instruction.  }
   \label{fig:method}
\end{figure}

\subsection{Dataset Utilization} %\textcolor{blue}{[Proposed Dataset following]}

% The conventional instruction-editing model usually consists of a text encoder and a diffusion model.  The text encoder, typically adopting the CLIP text encoder, is responsible for extracting the salient features from the instruction to guide the subsequent image editing process. Simultaneously, the diffusion model accesses these extracted features and diffuses the noised input image to align it with the provided instruction. 
We utilize our extensive model to enhance the active reasoning capability of the editing model. Concretely, we design a simple framework, which integrates a Multimodal Large Language Model (MLLM) into the diffusion model, as depicted in Figure~\ref{fig:method}. Diverging from previous methodologies that comprehend human intent solely through text, the MLLM enhances understanding by incorporating both the instruction and the input image. Formally, the instruction feature $h$ with human's intent can be formulated as
\begin{align}
\begin{aligned}
    h = & \;  \mathcal{F}(\vx_{input}, I),
\end{aligned}
\end{align}
where $\mathcal{F}$ is the MLLM. $h$ is the output of $\mathcal{F}$, containing the multimodal understanding of our instruction.

% Then we moved to the architecture of our method. The original stable diffusion model can be divided into two parts, the text model and the image generative model. The former convert text input to condition supervision information while the latter conduct image generation or editing. In this paper, to enhance the active reasoning capability of the editing model, we introduce Multimodal Large Language Model to the framework. As shown in Figure~\ref{fig:method}, we adopt a Multimodal LLM, LLava~\misscite{} in our method. Unlike previous methods that only take text information, it absorbs both the image and text instruction. So

% \begin{align}
% \begin{aligned}
%     h_{out} = & \;  \mathcal{F}(\vx_{input}, c_{T}),
% \end{aligned}
% \end{align}

% where $\mathcal{F}$ is the Multimodal LLM, it takes both input image $\vx_{input}$ and text instruction $c_T$ into consideration. $h_{llm}$ is the output hidden state of the model, containing the multimodal understanding of our instruction. 

% To inject $h_{llm}$ to the text model in stable diffusion model seamlessly, we abandoned several layers in the text model $\mathcal{T}$ and project $h_{llm}$ to suitable dimension by a projection layer $\gamma$. Therefore, the text condition will be 
% \begin{align}
% \begin{aligned}
%     c_{out} = \mathcal{T}(\gamma(h_{llm})),
% \end{aligned}
% \end{align}

Then we can inject $h$ seamlessly into the editing model. The image generative model can edit the input image under the supervision of $h$.
\begin{align}
\begin{aligned}
    \vx_{out} = \mathcal{M}(\vx_{input}, h),
\end{aligned}
\end{align}
where $\mathcal{M}$ is the image model, and $\vx_{out}$ is the corresponding output. 

Considering the large amounts of parameters in LLM, we fixed it when fine-tuning our model. With our ReasonPix2Pix dataset, the model is fine-tuned end-to-end.

\section{Experiment}
\label{sec:experiment}
\subsection{Implementation Details} 

We utilize GPT-3.5-turbo when generating our dataset. We adopt Stable Diffusion v1.5~\cite{rombach2022high} and LLaVA-7B-v1.5~\cite{liu2023visual} in our fine-tuning process. The images are resized to $256 \times 256$, and the base learning rate is $1 \times 10^{-4}$ during training~\cite{wang2019distributed,wang2017distributed,zhou2018distributed,zhou2019finite}. Other training configs are consistent with those in InstructPix2Pix~\cite{brooks2023instructpix2pix}.

We utilize the test data of V3Det to construct a benchmark by the data generation pipeline in Figure~\ref{fig:gen2}, with $1000$ images. Meanwhile, we record the selected category and the target new category, so we can formulate a straightforward instruction by multiple templates, \eg Turn A to B. Therefore, our test data consists of the input image and its caption, the ground-truth edited image and its caption, and the straightforward instruction and reasoning instruction, respectively. We evaluate our method, as well as previous methods on this test set.

% \subsection{Results}
\subsection{Qualitative Results}

\begin{figure}[pt]
   \begin{center}
   \includegraphics[width=1\linewidth]{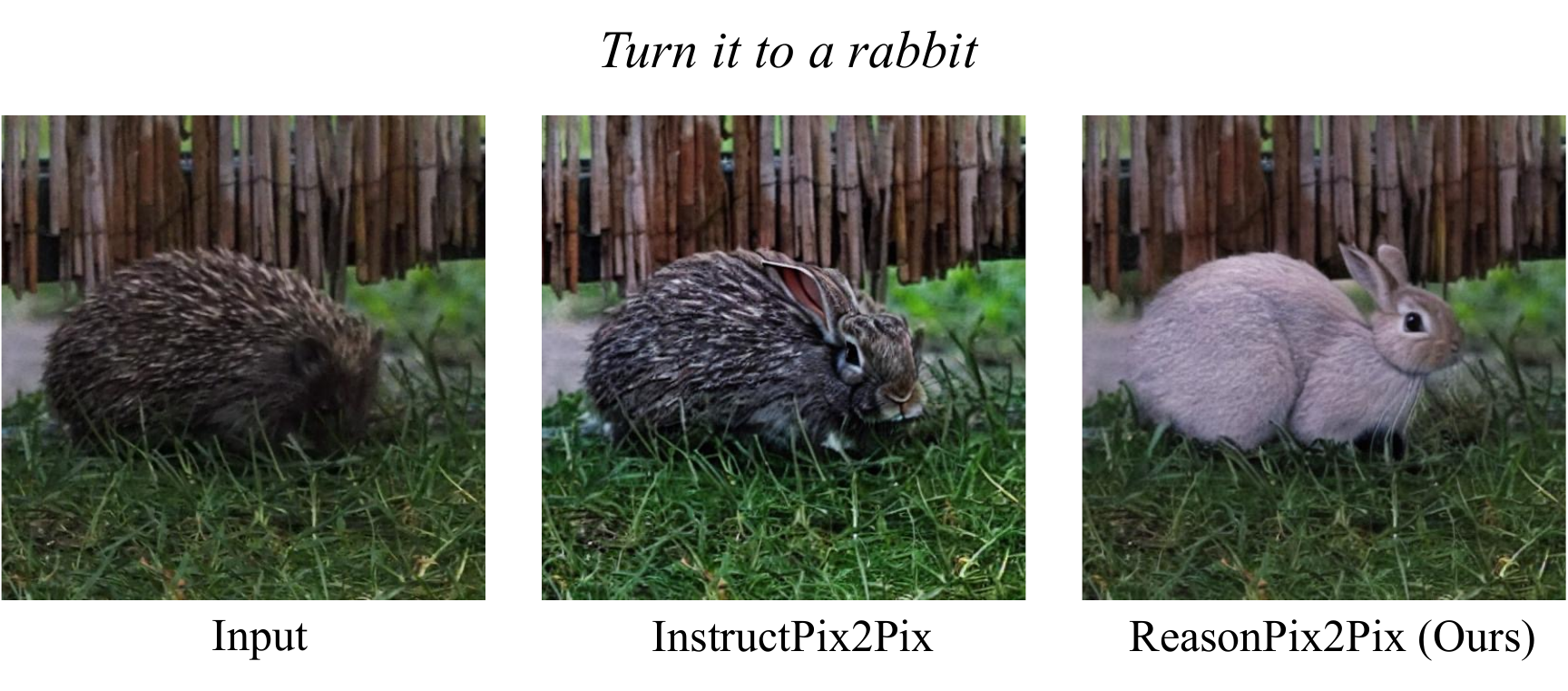}
   \end{center}
         \vspace{-10pt}

   \caption{Generated results with instruction \textit{``Turn it to a rabbit"}.}

   \label{fig:lake}
\end{figure}

\paragraph{Image Quality} Here we compare the performance of our method with previous methods under straightforward instructions. As in Figure~\ref{fig:lake}, InstructPix2Pix fails to turn the hedgehog in the image to a rabbit. Our method is able to convert these complicated categories, generating more vivid results. 

\begin{figure}[pt]
   \begin{center}
   \includegraphics[width=0.9\linewidth]{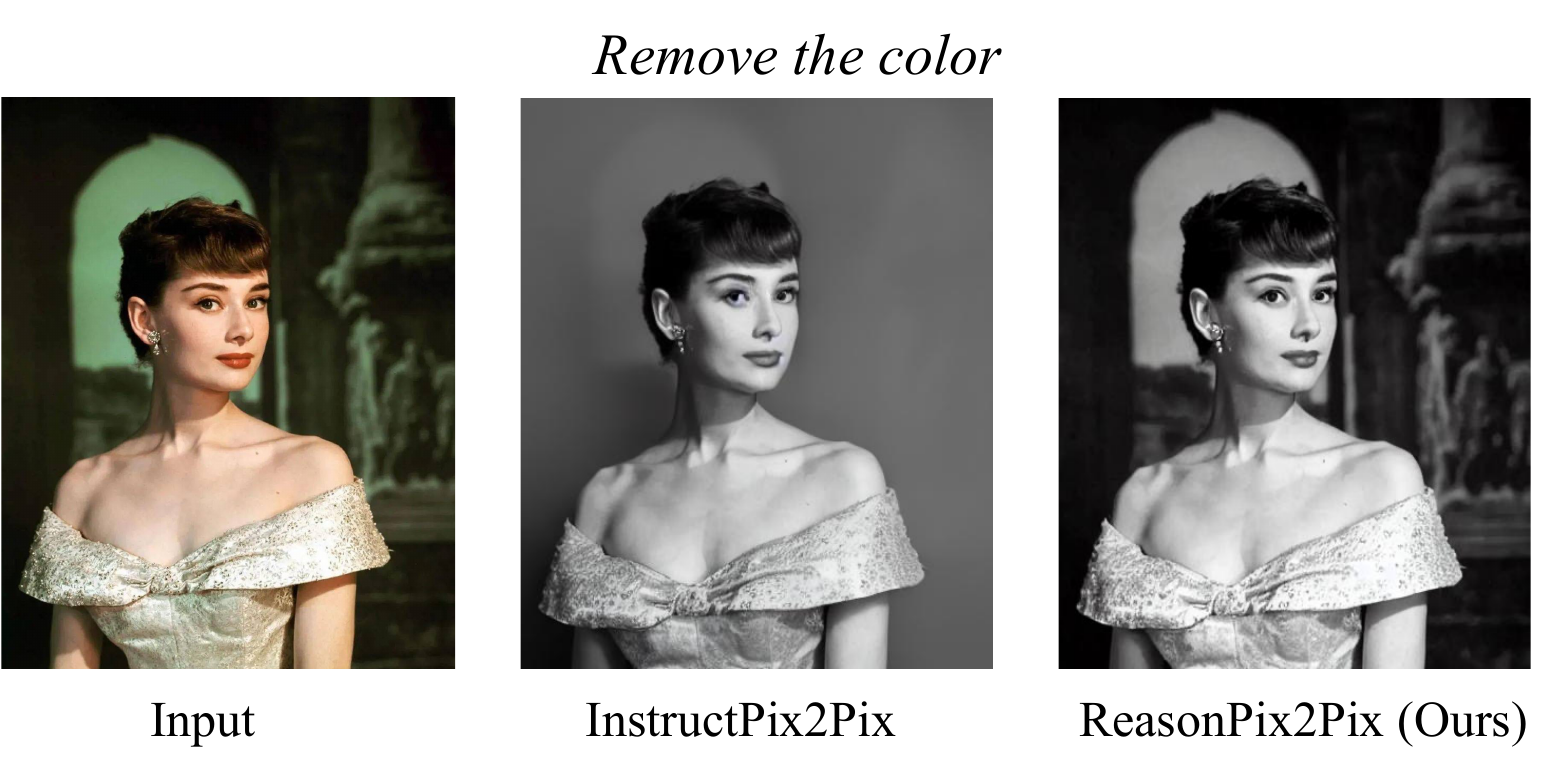}
   \end{center}
      \vspace{-10pt}

   \caption{Generated results with instruction \textit{``remove the color"}.}
   \label{fig:adl}
\end{figure}

\paragraph{Reasoning Ability} To compare the reasoning ability, first we start from relatively simple instruction. As shown in Figure~\ref{fig:adl}, when the instruction is \textit{``remove the color"}. Previous methods can to some extent, understand the instruction, but the generated results are not accurate. InstructPix2Pix follows the instruction to convert the image to black-and-white, but it also removes the background. On the contrary, our ReasonPix2Pix understands the instruction and gives adequate results. 

\begin{figure}[pt]
   \begin{center}
   \includegraphics[width=0.9\linewidth]{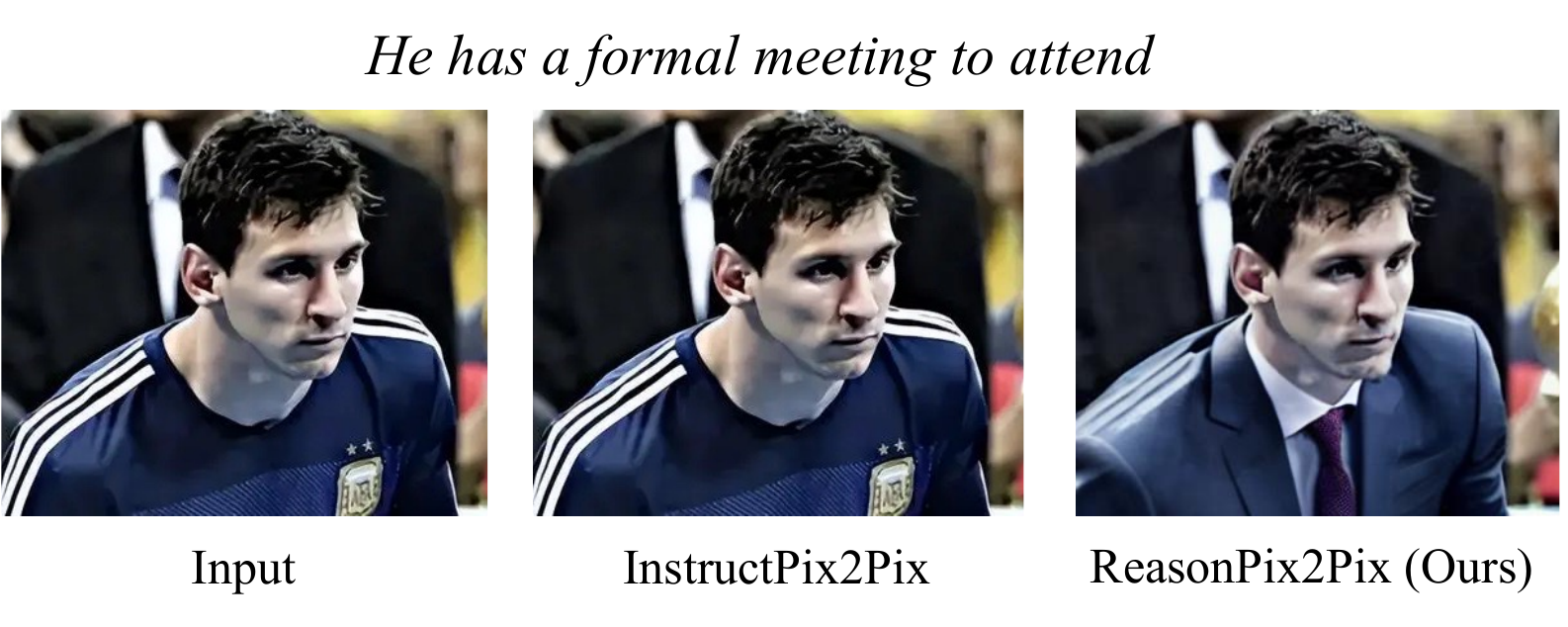}
   \end{center}
      \vspace{-10pt}

   \caption{Generated results with instruction \textit{``he has a formal meeting to attend"}.}
   \label{fig:scarlet}
\end{figure}
\begin{figure*}[htbp]
   \begin{center}
   \includegraphics[width=0.75\linewidth]{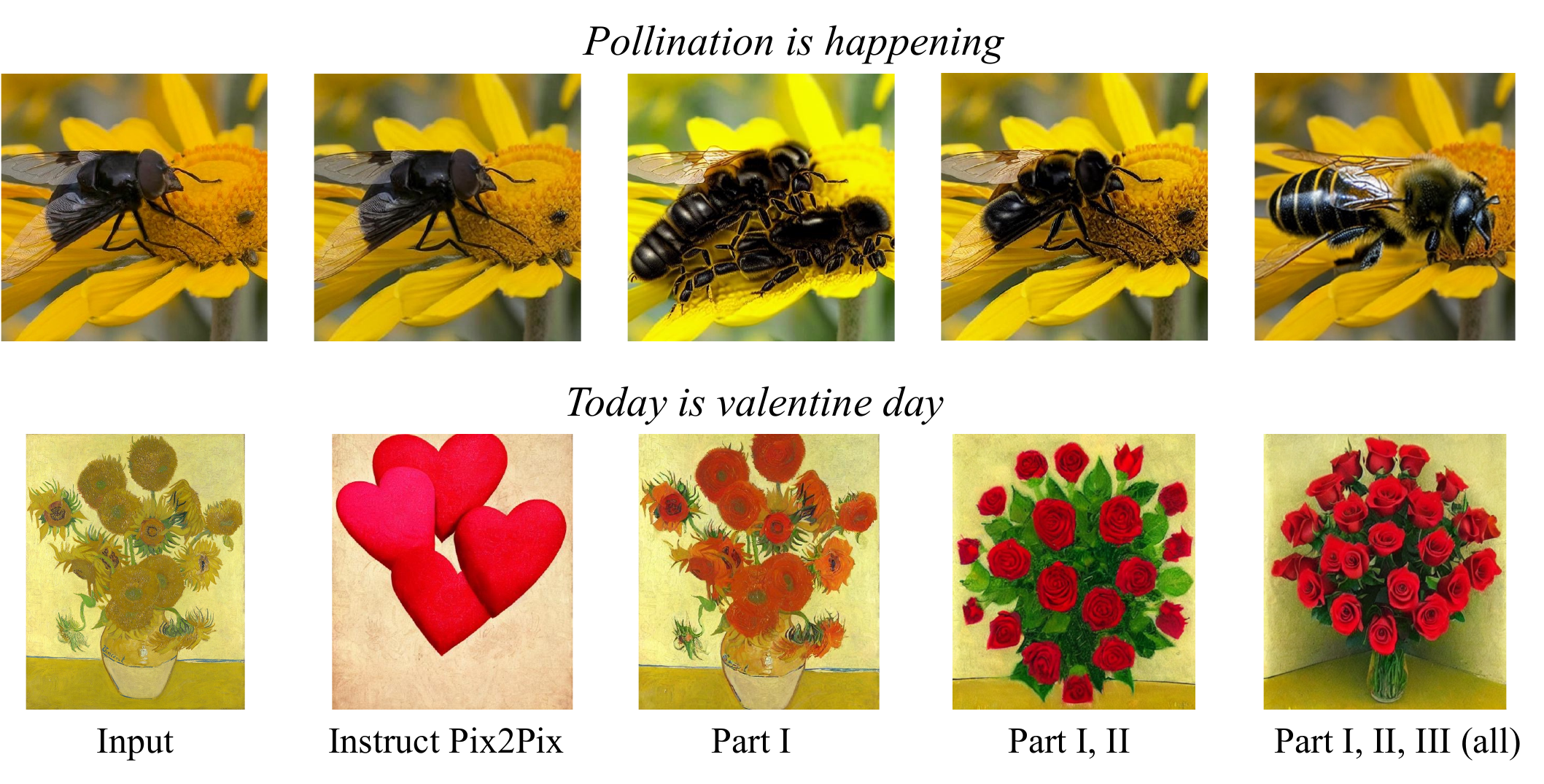}
   \end{center}
      \vspace{-10pt}
   \caption{Generated results with models trained on different parts of data. We show the input, InstructPix2Pix, our model trained on Part I data, Part I,II data, and Part I,II,III (all) data respectively.}
   \label{fig:data-ana}
\end{figure*}

\begin{table*}[]
\centering
\caption{Quantitative results on test set. $\downarrow$ / $\uparrow$ means the lower / higher the better.}
\resizebox{0.9\textwidth}{!}{
\begin{tabular}{l|ccccc|ccccc}
\toprule
\multirow{2}{*}{Method} & \multicolumn{5}{c|}{Direct Instruction} & \multicolumn{5}{c}{Reasoning Instruction}\\
                        &L1$\downarrow$  & L2$\downarrow$  & CLIP-I$\uparrow$ & DINO$\uparrow$  & CLIP-T$\uparrow$
                        &L1$\downarrow$  & L2$\downarrow$  & CLIP-I$\uparrow$ & DINO$\uparrow$  & CLIP-T$\uparrow$\\
\midrule
Null-text~\cite{mokady2023null} & 0.0931 & 0.0354 & 0.8542 & 0.8036 & 0.2479 & 0.2637 & 0.1165 & 0.6326 & 0.5249 & 0.1706 \\
InstructPix2Pix~\cite{brooks2023instructpix2pix} & 0.1265   & 0.0423   & 0.8042    & 0.7256  & 0.2465 & 0.2984                             & 0.1385 & 0.6034 & 0.5142 & 0.1629    \\  
MagicBrush~\cite{zhang2023magicbrush} & 0.0706   & 0.0247  & 0.9127  & 0.8745  & \textbf{0.2568} & 0.2239 & 0.0938 & 0.6755 & 0.6125 & 0.1941                    \\
EDICT~\cite{wallace2023edict} & 0.1149 & 0.0385 & 0.8137 & 0.7485 & 0.2490 & 0.2753 & 0.1296 & 0.6282 & 0.5526 & 0.1703\\
InstructDiffusion~\cite{geng2023instructdiffusion} & 0.0824  & 0.0295  & 0.8873 & 0.8461 & 0.2506 & 0.2145 & 0.0863 & 0.6904 & 0.6375 & 0.2046 \\
\midrule
ReasonPix2Pix (Ours) & \textbf{0.0646} & \textbf{0.0203} & \textbf{0.9246} & \textbf{0.8920} & 0.2553 & \textbf{0.1347} & \textbf{0.0476} & \textbf{0.7824} & \textbf{0.7216} & \textbf{0.2350} \\
\bottomrule
\end{tabular}
}
\label{tab:quant}
\end{table*}

\begin{table*}[]
\small
\centering
% \resizebox{\linewidth}{!}{
% \caption{Multi-choice comparison of four methods. The numbers represent the frequency of each method being chosen as the best for each aspect.}
\caption{User study of four methods. We sample $100$ images for each method, with direct and reasoning instruction, respectively. We report the frequency of images generated by each method being chosen as the best among four.}
\begin{tabular}{l|cccc}
\toprule & InstructPix2Pix~\cite{brooks2023instructpix2pix} & MagicBrush~\cite{zhang2023magicbrush} & InstructDiffusion~\cite{geng2023instructdiffusion} & ReasonPix2Pix (Ours)   \\
\midrule
Direct Instruction &  16 & 21 & 28 & \textbf{35}\\
Reasoning Instruction &   13 & 15 & 18 & \textbf{54}\\
\bottomrule
\end{tabular}
% }
% \vspace{1em}
% \caption{The results of Human Evaluation on comparing the edited images generated by four methods listed in the table. We randomly sampled 100 edited images generated by the aforementioned methods, and workers were asked two questions to select the best one based on \textit{Consistency} and \textit{Image Quality}, respectively. The numbers in the table represent the frequency of each method being chosen as the best.}
\label{tab:user_multichoice}
\vspace{-1em}
\end{table*}

Then we move to more complicated instructions. As demonstrated in Figure~\ref{fig:scarlet}, with an indirect instruction \textit{``he has a formal meeting to attend"}, previous InstructPix2Pix cannot tackle it, outputing the original image without any editing. Our method can understand the instruction, and ask him to wear formal clothes to attend the meeting.

\subsection{Quantitative Results}
Besides the qualitative results above, we also compare the quantitative metrics with previous methods in Table~\ref{tab:quant}, with direct instruction and reasoning instruction respectively. We report the L1 and L2 distance between the generated images and ground-truth images, and the cosine distance between their CLIP and DINO embeddings respectively. In addition, we also report CLIP-T, the cosine similarity between CLIP features of the target caption and the generated image. 
With traditional direct instructions, our method achieves competitive performance among previous methods, proving the quality of our generated images. When it comes to reasoning instruction that requires understanding, previous methods show degraded performance, but our method achieves remarkably higher results than other methods. 

Meantime, we also establish a user study to compare our method with previous methods. We random sample $100$ samples generated by different models respectively, and ask $5$ workers to evaluate them ($20$ per person). The workers are asked to pick out the best image among $4$ candidates. From Table~\ref{tab:user_multichoice}, with direct instruction, our method is mildly superior to previous methods. When the instructions become reasoning ones, the gap between our method and previous methods becomes larger.

\subsection{Analysis} 

\paragraph{Qualitative Results} We evaluate the effectiveness of the three parts of our dataset. Figure~\ref{fig:data-ana} demonstrates the results when training our method with merely Part I, Part I, and Part II, and the whole dataset respectively. We can observe that when confronting instructions that require reasoning, previous methods such as InstructPix2Pix are prone to edit nothing or produce unreasonable editing results. With Part I data, the model seems to understand the instructions, but it is still hard to provide an edited image. It is consistent with our proposal that with merely the images in InstructPix2Pix dataset, the editing ability of the model is still limited. On the other hand, when introducing Part II and Part III data sequentially, the editing results become increasingly better. With all the data in our dataset, the model is capable of understanding the instruction and producing the corresponding results. 

\begin{figure}[pt]
   \begin{center}
   \includegraphics[width=0.8\linewidth]{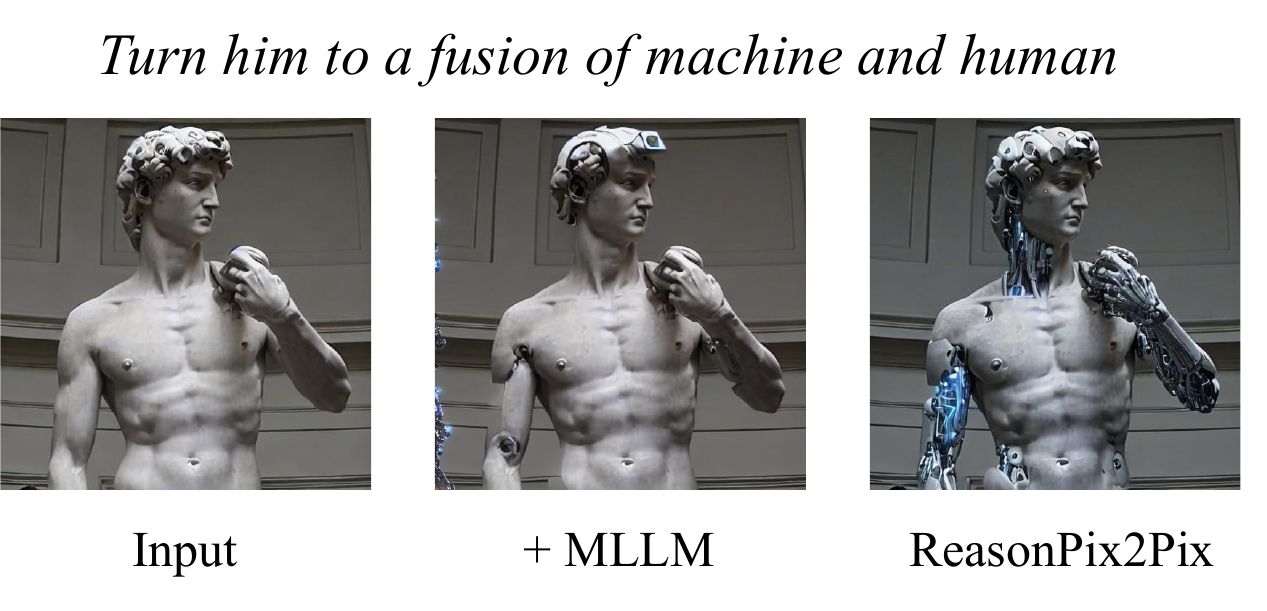}
   \end{center}
   \vspace{-10pt}
   \caption{Generated results with different models. We show the input, results generated without fine-tuning, and our method respectively.}
   \label{fig:method-ana}
\end{figure}

On the other hand, in our simple framework, we integrate Multi-modal Large Language model into the image editing model, which naturally has reasoning ability. Here, we compare the results of InstructPix2Pix, adding MLLM without fine-tuning, and our model that is fine-tuned on ReasonPix2Pix. Figure~\ref{fig:method-ana} shows that without fine-tuning, it is hard for the image editing model to take the output of MLLM. When fine-tuned on our dataset, the model is capable of understanding and editing. 

\begin{figure}[tbp]
    \centering
    \begin{subfigure}{0.45\linewidth}
        \includegraphics[width=\textwidth]{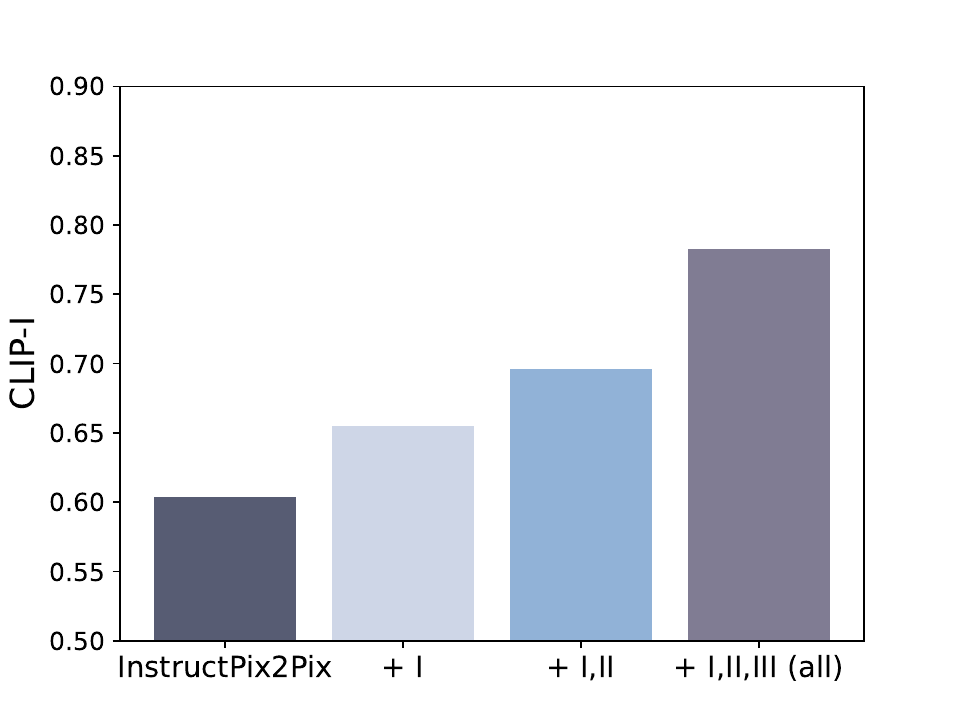}
        \caption{}
        \label{fig:quant-ablation1}
    \end{subfigure}
    \begin{subfigure}{0.45\linewidth}
        \includegraphics[width=\textwidth]{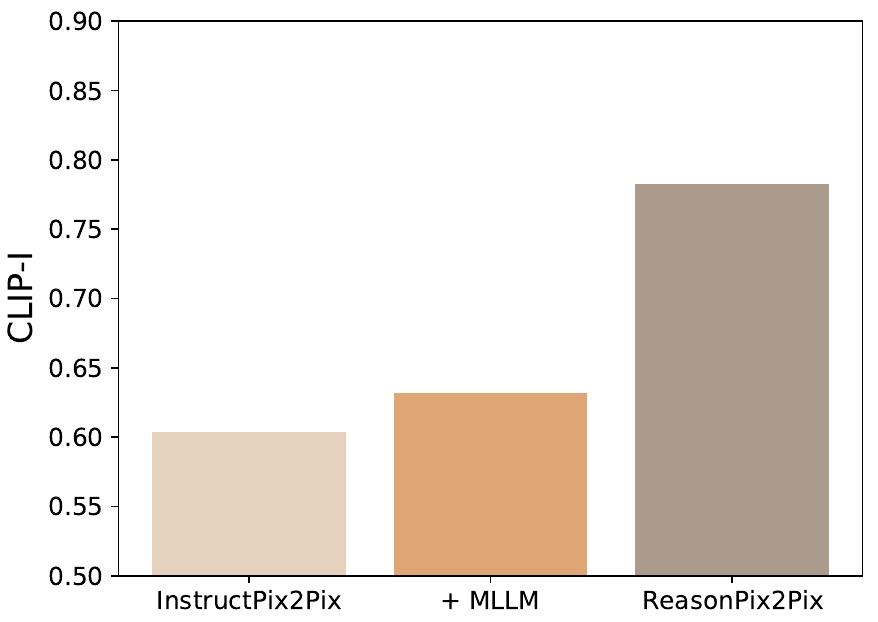}
        \caption{}
        \label{fig:quant-ablation2}
    \end{subfigure}
    \caption{Ablation Study. (a) CLIP-I results of InstructPix2Pix, our method with Part I, Part I,II and Part I,II,III (all) data. (b) CLIP-I results on InstructPix2Pix, zero-shot image editing with MLLM, and the model fine-tuned on our ReasonPix2Pix. }
    \vspace{-10pt}
\end{figure}

\paragraph{Quantitative Results} In Figure~\ref{fig:quant-ablation1} we compare the quantitative results. CLIP-I rises when we add Part I, II, and III data. Therefore, the three parts of our dataset are all indispensable. Meanwhile, as shown in Figure~\ref{fig:quant-ablation2}, MLLM brings about minor improvements, and our dataset obviously advances model performance. The quantitative results are again, consistent with our qualitative results. 

\paragraph{Comprehensive Understanding} 

\begin{figure}[pt]
   \begin{center}
   \includegraphics[width=0.8\linewidth]{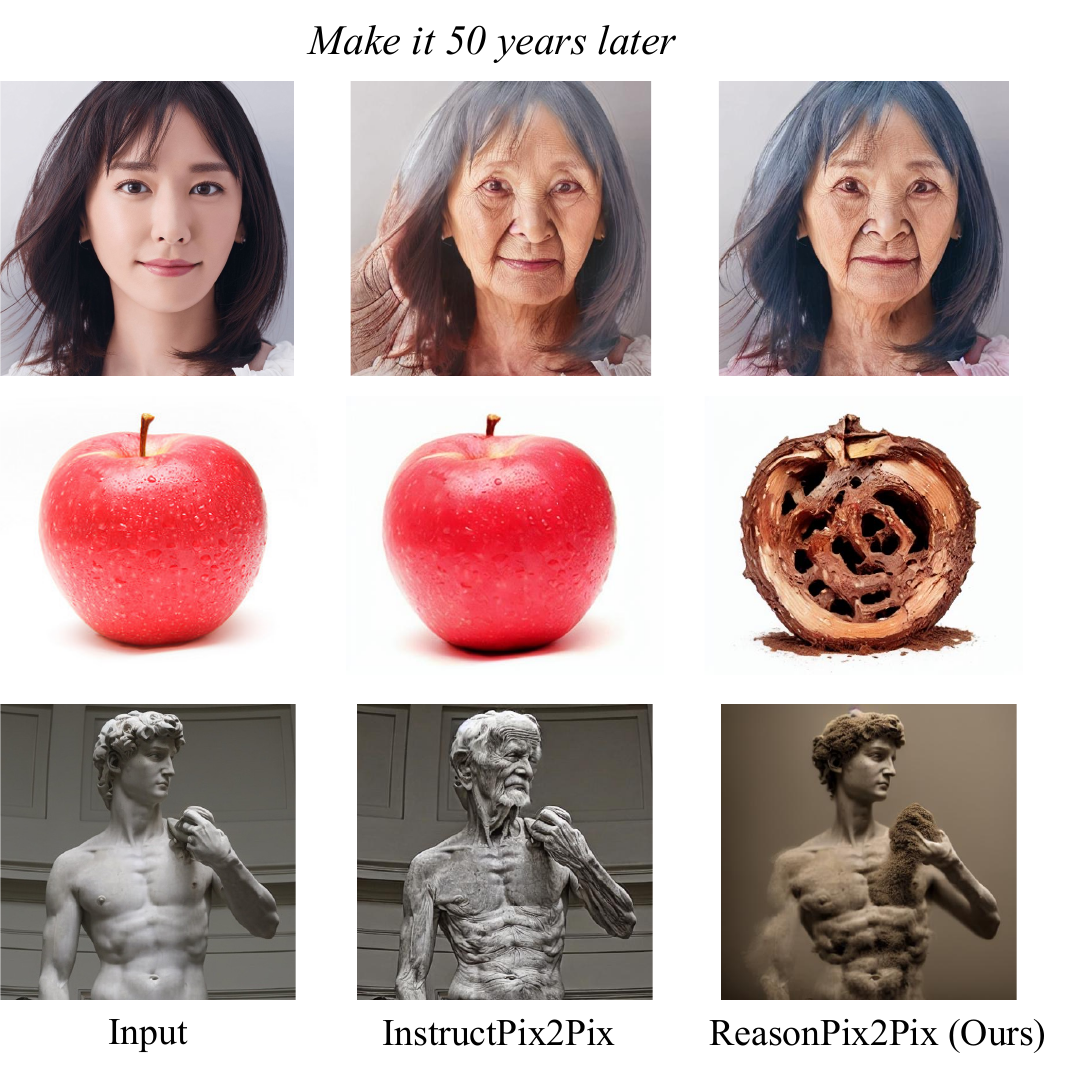}
   \end{center}
 \vspace{-10pt}
   \caption{Generated results with the instruction \textit{``make it 50 years later"}. Our model can have a comprehensive understanding of one instruction with different images, and provide adequate results.}
   \label{fig:50-years-ans}
\end{figure}

Finally, let us return the case in Sec.~\ref{sec:intro}, the instruction \textit{``make it 50 years later"}. As mentioned in Sec.~\ref{sec:intro}, previous methods cannot cope with some cases such as fruits. Meanwhile, understanding instruction is not a single-modal problem, a statue of a man will not become an old man after 50 years. With our framework and dataset, the model takes both image and instruction into consideration. As a result, it provides reasonable results according to different inputs. After 50 years, a young beautiful woman becomes an old woman, the apple turns into a rotted one, and the statue becomes a broken one with dust.

% \paragraph{Method} It will be insteresting to dig deeper into our method design. In our method, we take the output hidden state of LLM and project it to the input of text condition model, so how about taking the output text? On the other hand, is it train both lauguage model and image model in our framework? We conduct ablation study on our method and the results are compared in Figure~\ref{fig:method-ana}. First, from column 1 and 2, taking LLM output text is sometimes applicable (line 2), but it is unreliable (line 1). The failure case in line 1 can be attributed to too long text for the text model to handle. Meanwhile, training merely the image model can tackle relatively straightforward instructions \textit{"Turn him to a fusion of machine and human"}, but fails when confronting implicit instruction \textit{"Flower smells well"}. 

% At the same time, Figure~\ref{fig:quant-ablation2} prove that our method performs best from a quantitative perspective. Our designed method achieves the best results among all the variants, proving that our design is suitable for instruction editing tasks.  

% \begin{figure}[pt]
%    \begin{center}
%    \includegraphics[width=1\linewidth]{./images/method-ana.pdf}
%    \end{center}
%    \caption{Generated results with different models. We show the input, results generated by taking LLM output text as the text condition, training merely image model in our framework, and our method that trains both image and language model respectively.}
%    \label{fig:method-ana}
% \end{figure}

\subsection{Limitations}

% \paragraph{Dataset Size} First, our dataset size is still limited due to API costs. We have formulated a clear data generation pipeline in this paper. If needed, researchers can expand our dataset to more than $400,000$. 

% \paragraph{Ambiguity} One big challenge in generating reasoning instruction-based image editing datasets is that it is difficult to eliminate ambiguity. For example, for instruction 'She has a formal meeting to attend', she can wear a dress or suit, forming two different edited images. In other words, one reasoning instruction may be mapped to multiple edited images. We have alleviated ambiguity when asking GPT-3.5 to generate and select instructions, but there are still some ambiguous cases in our dataset. We consider it worth researching in future works. 
Our dataset size is still limited due to API costs. We have formulated a clear data generation pipeline in this paper. If needed, researchers can expand our dataset to more than $400,000$.

\section{Conclusion}
\label{sec:conclusion}

In this paper, we aim at enhancing the reasoning ability of editing models to make it more intelligent. Concretely, we introduce ReasonPix2Pix, a dedicated reasoning instruction editing dataset to inject reasoning ability to image editing. We fine-tune a simple framework on our proposed dataset. Extensive experiment results prove that our method achieves competitive results, no matter the instruction requires reasoning or not.

{
    \small
    \bibliographystyle{ieeenat_fullname}
    \bibliography{main}
}

% WARNING: do not forget to delete the supplementary pages from your submission 
% \input{sec/suppl}

\end{document}